\documentclass[sigconf, noacm]{acmart}

\makeatletter
\def\@ACM@checkaffil{
    \if@ACM@instpresent\else
    \ClassWarningNoLine{\@classname}{No institution present for an affiliation}%
    \fi
    \if@ACM@citypresent\else
    \ClassWarningNoLine{\@classname}{No city present for an affiliation}%
    \fi
    \if@ACM@countrypresent\else
        \ClassWarningNoLine{\@classname}{No country present for an affiliation}%
    \fi
}
\makeatother

\usepackage{amsmath,amsfonts}
\usepackage{natbib}
\usepackage{graphicx}
\usepackage{textcomp}
\usepackage{enumitem}
\usepackage{hyperref}
\usepackage{balance} 
\usepackage{xcolor}
\usepackage{outlines}

\usepackage{comment}
\usepackage{hhline}
\usepackage{caption}
\usepackage{subcaption}
\usepackage{amsmath}
\usepackage[ruled,vlined]{algorithm2e}
\usepackage{tikz}
\usetikzlibrary{positioning}
\usepackage{pgfplots}
\usepackage{pgfplotstable}
\pgfplotsset{compat=1.7}
\usepgfplotslibrary{groupplots}
\usepackage{multirow}

\usepackage{comment}

\usepackage{blindtext}

\usepackage{etoolbox}

\usepackage{colortbl}
\makeatletter
\patchcmd{\maketitle}{\@copyrightspace}{}{}{}
\makeatother

\def\BibTeX{{\rm B\kern-.05em{\sc i\kern-.025em b}\kern-.08em
    T\kern-.1667em\lower.7ex\hbox{E}\kern-.125emX}}

\makeatletter
\def\addlegendimage{\pgfplots@addlegendimage}
\makeatother

\title{A Study on Domain Generalization for \\Failure Detection through Human Reactions in HRI}

\acmYear{2024}
\acmConference[SS4HRI at HRI'24]{SS4HRI: Social Signal Modeling in Human-Robot Interaction}{March 11}{Boulder, Colorado, USA}

\makeatletter
\def\@copyrightspace{\relax}
\makeatother

\begin{document}

\author{Maria Teresa Parreira}
\affiliation{%
  \institution{Cornell Tech}
}
\author{Sukruth Gowdru Lingaraju}
\affiliation{%
  \institution{Cornell Tech}
}
\author{Adolfo Ramirez-Aristizabal}
\affiliation{%
  \institution{Accenture Labs}
}

\author{Manaswi Saha}
\affiliation{%
  \institution{Accenture Labs}
}
\author{Michael Kuniavsky}
\affiliation{%
  \institution{Accenture Labs}
}
\author{Wendy Ju}
\affiliation{%
  \institution{Cornell Tech}
}

\renewcommand{\shortauthors}{Parreira et al.}

\begin{abstract}

Machine learning models are commonly tested in-distribution (same dataset); performance almost always drops in out of distribution settings. For HRI research, the goal is often to develop generalized models. This makes domain generalization -- retaining performance in different settings -- a critical issue.  
In this study, we present a concise analysis of domain generalization in failure detection models trained on human facial expressions. Using two distinct datasets of humans reacting to videos where error occurs, one from a controlled lab setting and another collected online, we trained deep learning models on each dataset. When testing these models on the alternate dataset, we observed a significant performance drop. We reflect on the causes for the observed model behavior and leave recommendations. This work emphasizes the need for HRI research focusing on improving model robustness and real-life applicability.

\end{abstract}


\maketitle

\section{Introduction}
Machine learning is increasingly used in affective computing to help machines and robots make sense of the social signals from the people around them. In machine learning for affective computing, researchers automate social signal recognition through the development of models trained on human reaction datasets; to evaluate these models, it is common practice to test these models on \textit{unseen} data from the same original dataset, with cross-validation techniques. However, because the data used for testing these models is selected from the same dataset as the training data, there are often underlying dataset characteristics, such as demographic and contextual factors, which impact the performance of these models when applied to out-of distribution data applications \cite{zhang2019general}. This hinders the generalizability and practical application of such models.

\begin{figure*}
    \centering
    \includegraphics[width=0.95\textwidth]{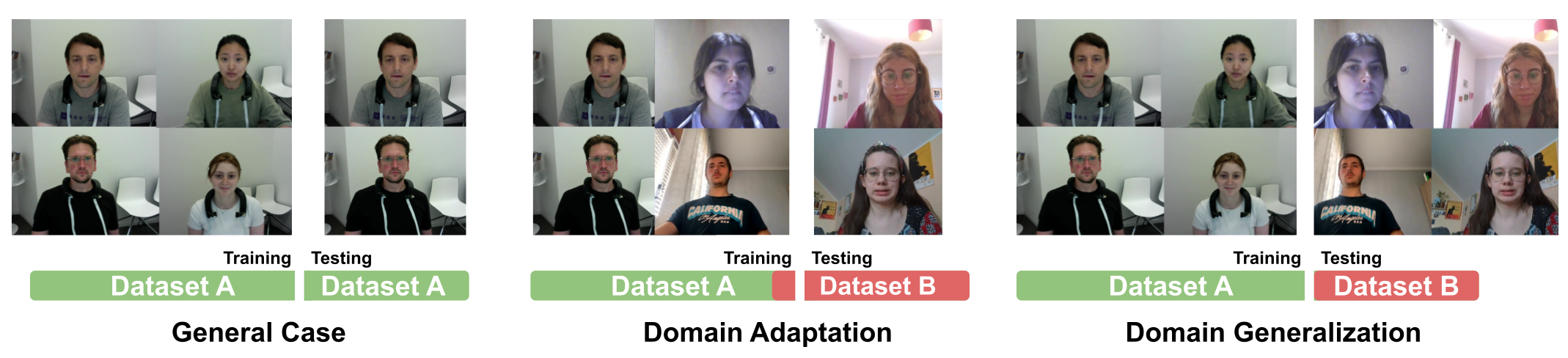}
    \caption{Illustration of concepts. In domain generalization, a model for Task X (e.g., failure detection through human reactions) is trained on a dataset A (e.g., from in-lab study) and deployed on an out-of-distribution dataset B (e.g., data collected in the wild). Domain adaptation includes some data from the testing dataset in the training data. Participants' image reproduced with consent.}
    \label{fig:domain}
\end{figure*}

In many robotics applications, researchers address this issue of limited model generalizability by collecting datasets for each novel application and training bespoke models. The fact that such models only perform well in the very specific conditions under which they were trained can be appropriate for well-constrained applications. However, a wider array of applications would be possible if social affect models were more broadly applicable. For this reason, some ML for affective computing researchers have been interested in how to adapt models to new contexts without significant loss of performance, \textit{transfer learning}. In particular, \textbf{domain generalization} consists of conserving the task but testing the model on a different target dataset/environment than on which the model was trained. Figure \ref{fig:domain} illustrates this concept.

\begin{figure}
    \centering
    \includegraphics[width=0.7\linewidth]{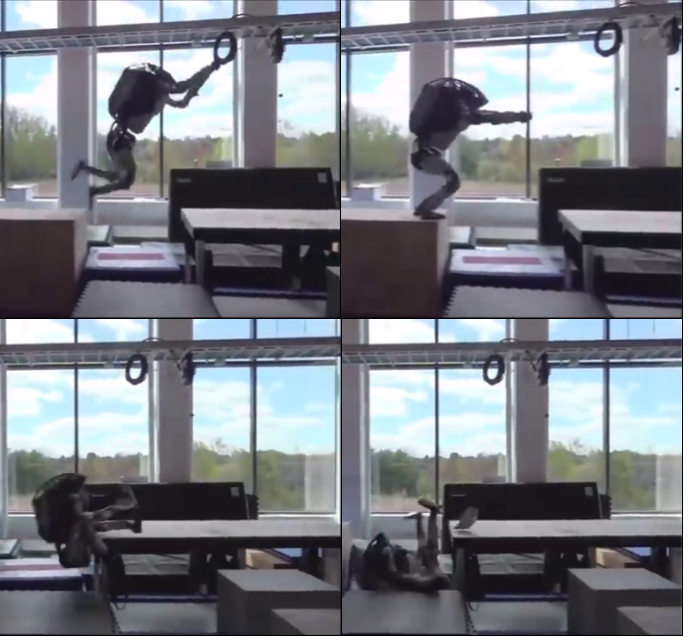}
    \caption{Example of stimulus video, displaying a robot failing a jump. These videos were shown to participants during data collection and included videos of human and robot error, as well as control videos. Video source included in the study repository.}
    \label{fig:stimvid}
\end{figure}

In this work, we present a brief analysis of domain generalization in models for failure detection through human reactions. This reaction from human bystanders can provide robots with a task-independent way to detect their own failures. 
For this application, we make use of two datasets of human reactions to videos of failures collected under different circumstances -- in-lab and online -- and cross-test models trained on each dataset, to evaluate the potential for domain generalization. We intend for this effort to contribute to a discussion on model robustness and generalizability in the field of affective HRI.

\section{Related Work}

\paragraph{\textbf{Social signal modelling for failure detection in HRI}}

Understanding human reactions, particularly to failure scenarios, is crucial for the development of socially intelligent and adaptive robots. Prior work has made significant progress in exploring how humans react to robot failure, and how we could explore these social cues for error detection \cite{stiber2023erroraware,stiber2022effective, kontogiorgos2020behavioural, bremers2023bystander}. Behaviors include body motion \cite{kontogiorgos2021systematic,2017trung,giuliani2015systematic}, gaze and and facial expressions \cite{kontogiorgos2021systematic,hwang2014reward,2020stiber,2010aronson}. Recently, \citet{bremers2023bystander} explored bystander reactions to failure and achieved high model performance in failure detection. Nonetheless, these and other HRI application models lack tests of generalization.


\paragraph{\textbf{Domain Generalization in Affective Computing}}

Transferability in affect computing and social robotics has been widely unexplored, with a few works revealing little promise for domain adaptation under the current ML frameworks for model development. For example, \citet{pineda2019domain} reviewed multiple works in automated facial computing (AFC) and found evidence of poor domain generalization, recommending caution when applying AFC in novel domains in which models have not been trained. \citet{suresh2023critically} also reports performance decreases in domain adaptation for facial expression recognition. Nonetheless, transferability is an issue that is increasingly being tackled in neighboring domains, such as robot manipulation, among other applications \cite{nan2019pepper,planamente2023domain}, with \citet{gao2023fer} reporting effective performance conservation in cross-domain techniques in affective computing. Greater transferability would help reduce the need for the time-consuming and challenging tasks of data collection and labeling, and model training for every new application, which hinder the rapid deployment and scalability of HRI systems.

\section{Study Design}
\label{sec:methods}

For collecting human reactions to observed errors, we followed the procedure in \citet{bremers2023bystander}. We designed an online user study (\textit{Online}) to collect visual responses to videos displaying failure. On a separate occasion, we ran an in-lab user study (\textit{Lab}), with the same protocol, to mimic data collection but in a different setting.

\subsection{Stimulus dataset}
\label{subsec:stimulus}

We selected a set of 30 stimulus videos for collecting human reactions. We included videos of human failures (10), robot failures (10) and control videos (10) (see Fig. \ref{fig:stimvid} for example). The videos' average length is $13.69\pm 7.77$ s. The full list can be consulted in the study repository \footnote{\url{https://github.com/mteresaparreira/baddata.git}}. 

\subsection{Experimental procedure}
\label{subsec:exp}

We followed the protocol in \citet{bremers2023bystander}. For the \textit{Online} study, participants were recruited online through Prolific. Flyers and word-of-mouth were used for the \textit{Lab} study. After giving informed consent, participants provided demographic information. After this, participants would see each stimulus video while their personal / the lab's laptop or computer webcam recorded their facial responses. Participants were not able to see their own image while the stimulus videos played, and the order of stimulus videos was randomized. Compensation was provided at rate of USD 15/hour for participants who completed the study (watched all 30 videos) and took less than 60 min. The full procedure took around 30 minutes to complete. 

\subsection{Participant demographics} 
\label{subsec:metrics}

We collected demographic information (age, gender, nationality and race/ethnicity). For the \textit{Online} dataset, the 29 participants took on average 33m55s$\pm$8m8s to complete the survey. Ages range from 19-37 ($25.4\pm 5.0$). 14 participants identify as female, 15 as male. The study included participants from 11 nationalities. Racial/ethnical distribution includes 18 Caucasian/White or Asian/White, 6 African /African American/Black, 3 Hispanic/Latino and 2 Asian/Asian American participants. For the \textit{Lab} dataset, the 30 participants took on average 20m14s$\pm$4m3s to complete the survey. Ages range from 21-78 ($37.73\pm 17.37$). 17 participants identify as female, 12 as male, and 1 as non-binary. The study included participants from 11 nationalities. Racial/ethnical distribution includes 17 Caucasian/White or Asian/White, 8 Asian/Asian American participants, 4 Hispanic/Latino and 1 African/African American/Black.

\section{Model development: human reactions to observed failures} 
\label{sec:model}

To test domain generalization in failure detection models through affect, and in line with prior work \cite{bremers2023bystander,suresh2023critically}, we took a pre-trained model (ResNet50) \cite{He2015resnet} and fine-tuned the final linear layer. ResNet50 is a convolutional neural network (CNN) that is 50 layers deep and was trained on ImageNet.

\begin{description}[align=left, leftmargin=0em, labelsep=0.2em, font=\textbf, itemsep=0em,parsep=0.3em]

\item[Action Space: ]
The reactions were labeled according to the type of stimulus video shown to participants. In videos where error occurs, only the moments following the exact failure moment were used as input data (and labeled \textit{1, reaction to failure}). Control videos -- where there is no error -- were used integrally, and labelled \textit{0, neutral state}.

\item[Final dataset: ] The videos were converted into static frames that were extracted at a 5, 15, and 30 frames-per-second (fps) rate and, similarly to what was reported in \citet{bremers2023bystander}, resized to 224x224 pixels. Post-processing and data augmentation were performed on each frame, according to ImageNet \cite{deng2009imagenet} statistics for data normalization. 

The total number of frames per dataset can be seen on table \ref{tab:frames}. Because all participants watched the same stimulus videos, data is balanced across participants.

\begin{table}[]
\centering
\caption{Number of samples (frames) per dataset.}
\label{tab:frames}
\begin{tabular}{l|llll}
\textbf{Train Dataset} & \textbf{FPS} & \textbf{Failure} & \textbf{Neutral} & \textbf{Total} \\
\multirow{3}{*}{\textit{Online}} & 5 & 18322 & 25010 & 43332 \\
 & 15 & 54598 & 74714 & 129312 \\
 & 30 & 109142 & 149288 & 258430 \\ \hline
\multirow{3}{*}{\textit{Lab}} & 5 & 18790 & 25620 & 44410 \\
 & 15 & 56000 & 76498 & 132498 \\
 & 30 & 111932 & 152843 & 264775
\end{tabular}
\end{table}

\item[Model Training: ] Models were trained and tested using Pytorch libraries. We tested two different approaches to model training. In the \textit{mixed participant} folds approach, the dataset was randomly split into train-validation-test splits independently of participants; in the \textit{non-mixed participant} folds approach, the folds were split based on participant data, and the subsets of data do not have overlapping participants. We used stratified K-fold for the \textit{mixed participant} approach, to ensure labels were balanced for each fold. For the \textit{non-mixed participant} approach, we used a rate of 15 fps for extracting the frames, as this was deemed via prior testing to ensure sufficient data with reasonable computation time.

We performed hyperparameter tuning on a 70-20 train-val split, with 5 cross-validation folds, with accuracy as the scoring metric, reserving $10$\% of the dataset as the test set (for the \textit{non-mixed} participant approach, this is a 20/21-6-3 participant split). The best candidate model of each type was picked based on macro-accuracy on the validation set.

\item[Evaluation Metrics: ] The models were evaluated based on the macro averages of the following metrics: accuracy, f1-score, precision and recall, and Cohen's Kappa. Cohen's Kappa is a proxy for data labeling noise \cite{cohenskappa}, through normalizing classification performance by the random chance of agreeability between existing data labels and predicted model labels. We report model performance across the five folds.

\item[System Implementation: ] Training was performed using a CPU processor 96GB of RAM, with GPU acceleration (48GB). 
\end{description}
\section{Results} \label{sec:results}

\subsection{Model performance}

The best performing hyperparameters are the following: for \textit{mixed} participants, optimizer was Stochastic Gradient Descent (SGD), momentum 0.9, across all frame rates (5, 15, 30 fps); for \textit{non-mixed} participants, optimizer was Adadelta, $\rho=0.9$, $\epsilon=1e-06$, also across all frame rates. Both approaches used a starting learning rate of $0.001$, with a step decay factor of 0.1 every 10 epochs. Models were trained for a total of 70, 50, and 30 epochs for the 5, 15, 30 fps datasets, respectfully, as performance on the train and validation datasets were deemed stable after these epochs.

Table \ref{tab:results} displays the results for a mixed-participant approach, including the metrics across the 5 folds. Table \ref{tab:resultsnm} shows the results for non-mixed participants, or participant-dependent folds.

\begin{table*}[]
\centering
\caption{Model performance. Results presented in $M\pm SD$ across 5 folds. Test results shown for each dataset (in-distribution and out-of-distribution). Acc: accuracy; prec: precision.}
\label{tab:results}
\begin{tabular}{l|lllllllll}
\textbf{\begin{tabular}[c]{@{}l@{}}Train \\ Dataset\end{tabular}} & \textbf{FPS} & \textbf{\begin{tabular}[c]{@{}l@{}}Test \\ Dataset\end{tabular}} & \textbf{Train Acc} & \textbf{Val Acc} & \textbf{Test Acc} & \textbf{Test Recall} & \textbf{Test Prec} & \textbf{Test F1} & \textbf{Test Kappa} \\
\multirow{6}{*}{\textit{Online}} & \multirow{2}{*}{5} & \textit{Online} &  $0.911\pm 0.002$ & $0.906\pm 0.007$  & $0.954\pm 0.014$   & $0.954\pm 0.015$ & $0.952\pm 0.013$ & $0.953\pm 0.014$  & $0.906\pm 0.028$ \\

 &  &\cellcolor[HTML]{EFEFEF} \textit{Lab} &\cellcolor[HTML]{EFEFEF} - &\cellcolor[HTML]{EFEFEF} - &\cellcolor[HTML]{EFEFEF} $0.435\pm 0.010$ &\cellcolor[HTML]{EFEFEF} $ 0.517\pm 0.027$ &\cellcolor[HTML]{EFEFEF} $0.505\pm 0.007$ &\cellcolor[HTML]{EFEFEF} $0.343\pm 0.025$  &\cellcolor[HTML]{EFEFEF} $0.008\pm 0.013$ \\
 & \multirow{2}{*}{15} & \textit{Online} & $0.963\pm 0.011$ & $0.967\pm 0.011$ & $0.995\pm 0.005$ & $0.995\pm 0.005$ & $0.995\pm 0.006$ & $0.995\pm 0.006$ & $0.990\pm 0.011$ \\
 &  &\cellcolor[HTML]{EFEFEF} \textit{Lab} &\cellcolor[HTML]{EFEFEF}  -  &\cellcolor[HTML]{EFEFEF} -  &\cellcolor[HTML]{EFEFEF} $0.442\pm 0.009$ &\cellcolor[HTML]{EFEFEF} $0.514\pm 0.009$ &\cellcolor[HTML]{EFEFEF} $0.504\pm 0.004$ &\cellcolor[HTML]{EFEFEF} $0.373\pm 0.029$ &\cellcolor[HTML]{EFEFEF} $0.008\pm 0.006$ \\
 & \multirow{2}{*}{30} & \textit{Online}& $0.980\pm 0.002$ & $0.978\pm 0.006$ & $0.999\pm 0.002$ & $0.999\pm 0.002$ & $0.999\pm 0.002$ & $0.999\pm 0.002$ & $0.997\pm 0.004$ \\
 &  &\cellcolor[HTML]{EFEFEF} \textit{Lab}&\cellcolor[HTML]{EFEFEF}  -  &\cellcolor[HTML]{EFEFEF} -  &\cellcolor[HTML]{EFEFEF} $0.452\pm 0.018$ &\cellcolor[HTML]{EFEFEF} $0.52\pm 0.012$ &\cellcolor[HTML]{EFEFEF} $0.51\pm 0.007$ &\cellcolor[HTML]{EFEFEF} $0.391\pm 0.053$ &\cellcolor[HTML]{EFEFEF} $0.018\pm 0.012$ \\ \hline
\multirow{6}{*}{\textit{Lab}} & \multirow{2}{*}{5} & \textit{Lab}& $0.854\pm 0.004$ & $0.848\pm 0.013$ & $0.892\pm 0.018$ & $0.892\pm 0.018$ & $0.885\pm 0.019$ & $0.888\pm 0.018$ & $0.776\pm 0.037$ \\
 &  & \cellcolor[HTML]{EFEFEF}\textit{Online}&\cellcolor[HTML]{EFEFEF}  -  & \cellcolor[HTML]{EFEFEF}-  & \cellcolor[HTML]{EFEFEF}$0.571\pm 0.006$ &\cellcolor[HTML]{EFEFEF} $0.541\pm 0.007$ &\cellcolor[HTML]{EFEFEF} $0.527\pm 0.005$ & \cellcolor[HTML]{EFEFEF}$0.504\pm 0.015$ &\cellcolor[HTML]{EFEFEF} $0.059\pm 0.011$ \\
 & \multirow{2}{*}{15} & \textit{Lab}& $0.952\pm 0.012$ & $0.955\pm 0.01$ & $0.991\pm 0.006$ & $0.991\pm 0.006$ & $0.99\pm 0.007$ & $0.99\pm 0.006$ & $0.981\pm 0.013$ \\
 &  &\cellcolor[HTML]{EFEFEF} \textit{Online} &\cellcolor[HTML]{EFEFEF}  -  &\cellcolor[HTML]{EFEFEF} -  &\cellcolor[HTML]{EFEFEF} $0.548\pm 0.02$ &\cellcolor[HTML]{EFEFEF} $0.518\pm 0.009$ & \cellcolor[HTML]{EFEFEF}$0.513\pm 0.005$ & \cellcolor[HTML]{EFEFEF}$0.496\pm 0.009$ &\cellcolor[HTML]{EFEFEF} $0.027\pm 0.011$ \\
 & \multirow{2}{*}{30} & \textit{Lab} & $0.964\pm 0.004$ & $0.967\pm 0.003$ & $0.997\pm 0.001$ & $0.997\pm 0.001$ & $0.997\pm 0.001$ & $0.997\pm 0.001$ & $0.993\pm 0.002$ \\
 &  &\cellcolor[HTML]{EFEFEF} \textit{Online}&\cellcolor[HTML]{EFEFEF}  -  &\cellcolor[HTML]{EFEFEF} -  &\cellcolor[HTML]{EFEFEF} $0.54\pm 0.005$ &\cellcolor[HTML]{EFEFEF} $0.513\pm 0.007$ &\cellcolor[HTML]{EFEFEF} $0.512\pm 0.006$ &\cellcolor[HTML]{EFEFEF} $0.506\pm 0.007$ &\cellcolor[HTML]{EFEFEF} $0.024\pm 0.012$ \\
\end{tabular}
\end{table*}

\begin{table*}[]
\centering
\caption{Model performance for non-mixed participants approach. Results presented in $M\pm SD$ across 5 folds. Test results shown for each dataset (in-distribution and out-of-distribution). Acc: accuracy; prec: precision.}
\label{tab:resultsnm}
\begin{tabular}{l|lllllllll}
\textbf{\begin{tabular}[c]{@{}l@{}}Train \\ Dataset\end{tabular}} & \textbf{FPS} & \textbf{\begin{tabular}[c]{@{}l@{}}Test \\ Dataset\end{tabular}} & \textbf{Train Acc} & \textbf{Val Acc} & \textbf{Test Acc} & \textbf{Test Recall} & \textbf{Test Prec} & \textbf{Test F1} & \textbf{Test Kappa} \\
\multirow{2}{*}{\textit{Online}} & \multirow{2}{*}{15} & \textit{Online} &  $0.900\pm 0.004$ & $0.500\pm 0.017$ & $0.481\pm 0.015$ & $0.493\pm 0.014$ & $0.493\pm 0.013$ & $0.477\pm 0.013$ & $-0.014\pm 0.025$ \\
 &  &\cellcolor[HTML]{EFEFEF} \textit{Lab} &\cellcolor[HTML]{EFEFEF}  -  &\cellcolor[HTML]{EFEFEF} -  &\cellcolor[HTML]{EFEFEF} $0.431\pm 0.009$ & \cellcolor[HTML]{EFEFEF}$0.495\pm 0.162$ & \cellcolor[HTML]{EFEFEF}$0.5\pm 0.007$ & \cellcolor[HTML]{EFEFEF}$0.335\pm 0.044$ & \cellcolor[HTML]{EFEFEF}$0.001\pm 0.012$ \\ \hline
\multirow{2}{*}{\textit{Lab}} & \multirow{2}{*}{15} & \textit{Lab}& $0.846\pm 0.003$ & $0.544\pm 0.053$ & $0.547\pm 0.05$ & $0.549\pm 0.041$ & $0.546\pm 0.04$ & $0.536\pm 0.045$ & $0.092\pm 0.078$ \\
 &  &\cellcolor[HTML]{EFEFEF} \textit{Online}& \cellcolor[HTML]{EFEFEF} -  & \cellcolor[HTML]{EFEFEF}-  & \cellcolor[HTML]{EFEFEF}$0.557\pm 0.022$ &\cellcolor[HTML]{EFEFEF} $0.527\pm 0.02$ & \cellcolor[HTML]{EFEFEF}$0.517\pm 0.017$ &\cellcolor[HTML]{EFEFEF} $0.478\pm 0.057$ &\cellcolor[HTML]{EFEFEF} $0.036\pm 0.037$ \\ 

\end{tabular}
\end{table*}
\section{Discussion} \label{sec:discussion}

This work takes an important step in the understanding of the realistic potential for error detection through human reactions. It aims to be \textit{in no way exhaustive}, but rather a study that builds on prior work on domain generalization and brings up conversations on meta-methodology in the field of HRI. 

In this study, we implemented failure detection models based on two different datasets, while also evaluating differences in the extraction frequency (fps) of the data. In the \textit{mixed} participants approach, across both datasets, we observe that more data -- more fps -- leads to better training performance. Interestingly, the dataset collected in-lab, in spite of being more homogeneous, leads to lower training performances. 

For the mixed participants approach, we observe high performance in-distribution but low out-distribution performance, in all cases. The observed performance drop is not surprising due to inherent differences in data distribution between the two datasets. Models trained on one dataset may not generalize well to unseen data from another domain, as they may fail to capture the variations present in diverse real-world scenarios. Factors such as lighting conditions, camera angles, or cultural differences can significantly impact model performance, highlighting the importance of robustness testing across varied contexts. Nonetheless, performance on the \textit{in-lab} dataset-trained model seems to retain better out-distribution, with lower performance drops. 
This could be because successful training on the heterogeneity of the \textit{online} dataset requires a much larger sample size. Evidence for this can be seen in how recall is the highest performance metric when testing the \textit{online}-trained models on the \textit{lab} dataset. Another factor to consider is that in-lab data collection tends to engage participants more, which may lead to more visible external reactions when compared to online data collection.

The non-mixed participants approach shows low in-domain generalizability for new participants, with lower performances than the mixed approach, as well as low out-domain generalizability. Again, this is not surprising. People's reactions to failure are diverse, varying both in the behavior exhibited (e.g people may laugh or they may frown) \cite{kontogiorgos2021systematic} and in the degree of this behavior (e.g. smile versus out-lout laugh) \cite{2020stiber}. This within-participant variation paired with the across-participant differences demands datasets with a larger number of participant examples, especially if the intended use case for the system is generalization to new users.

Incorporating more diverse and representative data during model training, along with robust feature engineering and regularization techniques, can help enhance the model's ability to generalize across different scenarios. HRI research commonly uses feature extraction techniques such as facial units activation as a proxy for externally displayed behavior. While this combats some of the heterogeneity in data, by fixing facial points where activation is monitored, it comes at the cost of computation time for online applications, error propagation, and potential information loss by discretization of the environment. 

Researchers must make clear what the intended use case is. Are we building generalizable, real-world applicable tools? Or exploring controlled scenarios? Is this a single-user tool, or multiple users? For example, single-user tools might consider different concepts of ``generalization'', such as attempting to get good performances on a small batch of participants with as little training data as possible from each participant.

Ultimately, efforts to improve model robustness and real-life applicability in HRI research need a holistic approach that considers the complexities of human behavior and interaction across varied contexts.
   


\subsection{Limitations and Future Work}
\label{sec:limit}

The intention with this work is to test and illustrate a case of (lack of) domain generalization in a HRI setting. The scope of the paper is limited and the intention is to start a metamethodological discussion on the field of applied machine learning for robot interaction systems. Nonetheless, we recognize and enumerate limitations and opportunities to expand this analysis. 
While we only explored the fine-tuning of ResNet50, there are other pre-trained models that could have been explored, as well as other architectures for the final layer(s) that could have been used. While we made efforts to minimize data imbalances, another approach could have been to randomly sample the datasets and ensure an exact balance of failure and neutral participant responses. It would also be interesting to train models on both datasets simultaneously and observe if that improves the non-mixed participants approach; or test directly the effectiveness of strategies for cross-domain generalization \cite{gao2023fer}. Finally, while we explore only the visual modality of data, there are opportunities to explore domain generalization across modalities. 

We do not expand the discussion towards alternatives for real-time detection. For example, models might detect one error instance in a few but not all frames in the data for that instance. While that would be successful error detection, performance metrics as they are presented in this paper would not reflect that.
We invite researchers to reflect on the topics brought up by this work when developing and reporting on machine learning models for HRI applications.

\bibliographystyle{ACM-Reference-Format}
\balance
\bibliography{newbib.bib,bremers.bib}
\end{document}